\documentclass{article}
\usepackage{arxiv_style,times}
\usepackage{amsmath,amsfonts,bm}

\def\eqref#1{equation~\ref{#1}}
\def\1{\bm{1}}

\DeclareMathAlphabet{\mathsfit}{\encodingdefault}{\sfdefault}{m}{sl}
\SetMathAlphabet{\mathsfit}{bold}{\encodingdefault}{\sfdefault}{bx}{n}

\newcommand{\E}{\mathbb{E}}

\newcommand{\Var}{\mathrm{Var}}

\usepackage{hyperref}
\usepackage{url}
\usepackage{booktabs}
\usepackage{amsmath,amssymb,amsthm}
\usepackage{graphicx}
\usepackage{subcaption}
\usepackage{xcolor}
\usepackage{microtype}
\usepackage{placeins}
\hypersetup{colorlinks=true,linkcolor=black,citecolor=black,urlcolor=blue}

\newtheorem{proposition}{Proposition}
\newtheorem{theorem}{Theorem}
\newtheorem{corollary}{Corollary}
\newcommand{\Valias}{V_{\mathrm{alias}}}
\newcommand{\Vproc}{V_{\mathrm{proc}}}

\title{Why Does the Future Branch?\\Identifiable Closure Tests for Stochastic Physical World Models}

\author{Yibin Dong\\
Shandong University}

\begin{document}
\maketitle
\thispagestyle{plain}
\pagestyle{plain}

\begin{abstract}
A calibrated stochastic world model can reveal how uncertain a future is
without revealing \emph{why} it branches. The same conditional future law can
arise because an observation aliases physical states or because dynamics remain
random after the declared full state is fixed. We prove that ordinary
transitions cannot identify these two sources, even for a perfect probabilistic
predictor. \emph{ClosurePairs} makes them identifiable by crossing compatible
microstates with repeated exogenous disturbances and estimating state, noise,
and state--noise interaction variance. The central consequence is operational:
under finite hierarchical sampling, forecast difficulty governs the useful
compute scale, while the alias/process composition provides complementary
information about its direction---resolving the current state or sampling
future randomness. ClosurePairs recovers
source attribution at unchanged likelihood, reduces equal-budget decomposition
error in a nonlinear interaction benchmark, and supports observation-only
routing. On exact-marginal MetaWorld twins, an output-only allocator is at chance
while a Closure-supervised probe on frozen JEPA-WM features routes
$89.8$--$100\%$. In an independent ManiSkill PushCube confirmation, a
stochastic RSSM's outputs and latents remain at chance, whereas an RGB-only
Closure probe routes $100\%$ under both ID and geometry/camera OOD over five
seeds, matching direct allocation rather than exceeding it. Across five unseen
allocation menus, the same Closure probe routes $92.5\%/90.4\%$ ID/OOD with no
new oracle labels, versus $37.9\%/32.9\%$ for a frozen direct allocator.
ClosurePairs is therefore an identifiable, reusable mechanism
target that cannot be recovered from forecast quality alone.
\end{abstract}

\section{Introduction}

World models support prediction and planning by representing a distribution
over possible futures \citep{ha2018world,hafner2025dreamer}. In physical settings,
multiple futures are not an implementation nuisance: partial observation hides
velocities and material parameters, coarse-graining creates memory, and thermal
or randomized dynamics remain stochastic. A model should therefore predict a
distribution rather than a single trajectory. Yet a predictive distribution
does not explain its own width.

Consider two action-conditioned environments. In the first, the observation
omits a microstate variable that deterministically controls the future. In the
second, the observation is a complete state but fresh process noise controls the
future. The two environments can induce exactly the same $p(y\mid z,a)$. Every
proper scoring rule gives an optimal predictor the same expected score in both
\citep{gneiting2007proper}, although the correct intervention differs: acquire a
finer observation in the first environment and preserve stochastic branches in
the second. This is not the standard epistemic--aleatoric distinction
\citep{kendall2017uncertainties}; it concerns two sources inside the environment's
conditional randomness, relative to a declared state and intervention boundary.

We study this ambiguity as an \emph{identification problem}. Let $X$ be a full
simulator state, $Z=g_r(X)$ its observation at resolution $r$, $A$ an action,
$E$ an exogenous disturbance, and $Y=F(X,A,E)$ a future outcome. The law of total
variance splits forecast uncertainty into variation of the microstate-conditional
mean and residual variation at fixed microstate. We call these state aliasing and
process stochasticity. Ordinary samples $(Z,A,Y)$ reveal only their sum.

ClosurePairs augments evaluation data with two paired interventions: vary
compatible microstates while holding $(Z,A)$ fixed, and repeat disturbances while
holding $(X,A)$ fixed. In a simulator, crossing the same sampled disturbances
with every compatible microstate enables a functional-ANOVA decomposition
\citep{sobol1993sensitivity}. Crucially, shared random numbers alone are
insufficient in nonlinear systems: a state--noise interaction must be estimated
and assigned to the process component. When a simulator cannot replay a
disturbance, independent nested repeats remain sufficient.

\paragraph{Main conclusion.}
Forecast quality measures the \emph{amount} of predictive difficulty, but not
its physical source. Under finite hierarchical sampling, difficulty governs the
useful compute scale, while the alias/process composition provides complementary
information about its direction: resolve state aliasing or preserve process
branches. Our contributions establish this conclusion in three steps:

\begin{itemize}
  \item We prove observational non-identifiability of uncertainty-source
  attribution, define the target relative to a declared state boundary, and
  show that the ambiguity survives perfect likelihood and calibration.
  \item We turn classical nested-repeat and two-way random-effects estimators
  into an explicit world-model evaluation contract. The crossed protocol
  identifies state, noise, and interaction effects; its labels can be distilled
  into a router that receives no paired futures at test time.
  \item We show analytically and empirically that, under finite hierarchical
  sampling, difficulty governs useful compute scale while Closure provides a
  complementary state-versus-noise allocation signal. Equivalent Gaussian
  systems, an equal-budget nonlinear benchmark, exact-marginal MetaWorld twins,
  and a stochastic-RSSM ManiSkill confirmation provide complementary tests.
\end{itemize}

Our novelty claim is deliberately limited: the variance identities, common
random numbers, functional ANOVA, and random-effects estimators are classical.
The contribution is their combination into a world-model evaluation contract
that specifies an intervention boundary, prevents nonlinear interaction from
being misassigned, and connects source attribution to sensing and branching
decisions. We do not claim a new variance decomposition, stochastic-closure
architecture, or uncertainty taxonomy.

\section{Problem setup}

Fix an observation context and action $(Z=z,A=a)$. Compatible microstates follow
$X\sim q(\cdot\mid z,a)$ and exogenous disturbances $E\sim p_E$ are independent
under the intervention. For a scalar, square-integrable future
$Y=F(X,a,E)$, define
\begin{align}
\Valias(z,a)
  &= \Var_X\!\left[\E_E[Y\mid X,a]\mid z,a\right], \\
\Vproc(z,a)
  &= \E_X\!\left[\Var_E(Y\mid X,a)\mid z,a\right].
\end{align}
The conditional law of total variance gives
\begin{equation}
\Var(Y\mid z,a)=\Valias(z,a)+\Vproc(z,a). \label{eq:total}
\end{equation}
State aliasing is reducible by resolving information about $X$. Process
stochasticity remains when the declared $X$ is fixed. The qualifier
``declared'' matters: a positive process term is not a claim of fundamental
randomness, only irreducibility relative to the modeled state, time step, and
randomization interface.

For vector outcomes, the same definitions yield positive-semidefinite covariance
components. Our experiments use scalar physical summaries so that attribution,
confidence intervals, and sensing costs are directly interpretable.

\section{What observational prediction cannot identify}

\begin{proposition}[Observational non-identifiability]
For any $V>0$ there is a continuum of controlled systems with the same observed
kernel $p(Y\mid Z,A)$ and different $(\Valias,\Vproc)$.
No estimator based only on i.i.d. $(Z,A,Y)$ can consistently recover both
components over this class.
\label{prop:nonid}
\end{proposition}

\begin{corollary}[Predictive-objective non-identifiability]
Let a population training or evaluation criterion depend only on the observed
law $P(Z,A,Y)$ and on a predictor's conditional law $Q(Y\mid Z,A)$. No rule
based only on that criterion can uniformly recover $(\Valias,\Vproc)$ over the
observationally equivalent family in Proposition~\ref{prop:nonid}. This includes
likelihood, proper scoring rules such as CRPS, calibration criteria, and
distributional objectives such as diffusion score matching when they receive no
additional state, intervention, or structural information.
\label{cor:objective-nonid}
\end{corollary}

The observed law, the criterion, and its set of population-optimal predictive
distributions are identical throughout the family, while the target split
changes with $\lambda$. The qualifier ``based only on'' is essential: the result
does not rule out identification from ClosurePairs interventions, a declared
full state, or additional structural assumptions.

For $\lambda\in[0,1]$, let
\begin{equation}
Y=\mu(Z,A)+\sqrt{\lambda V}\,U+\sqrt{(1-\lambda)V}\,E,
\label{eq:family}
\end{equation}
where $U,E$ are independent standard Gaussians and $U$ is hidden by $Z$. Every
member has $Y\mid Z,A\sim\mathcal N(\mu(Z,A),V)$, while
$(\Valias,\Vproc)=(\lambda V,(1-\lambda)V)$. The data distribution of any
observational estimator is therefore identical at parameter values with
different targets. Appendix~\ref{app:proofs} gives the formal proof.

\begin{figure}[t]
  \centering
  \includegraphics[width=\linewidth]{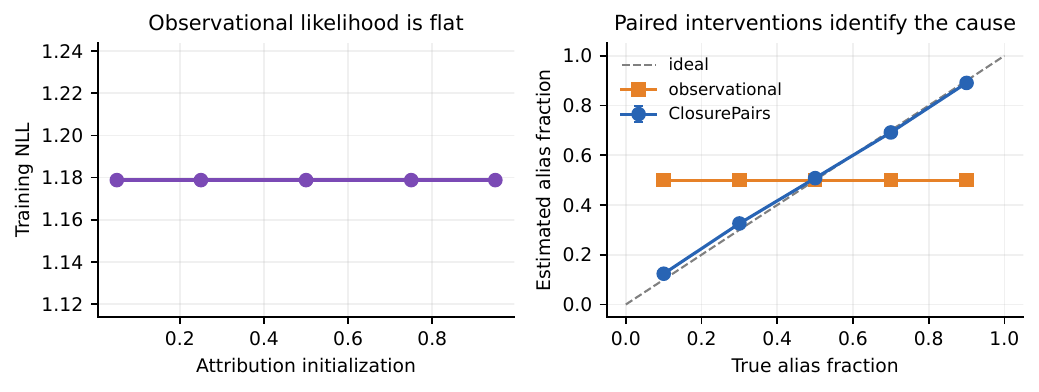}
  \caption{\textbf{Forecast equivalence does not imply mechanism equivalence.}
  Left: five neural attributions produce identical likelihood. Right: across
  observationally equivalent systems, the observational split stays at its
  arbitrary initialization, whereas ClosurePairs recovers the true split.
  Error bars are standard deviations over three seeds.}
  \label{fig:identifiability}
\end{figure}

Figure~\ref{fig:identifiability} shows the learned consequence. A neural
Gaussian predictor has a likelihood head for total variance and a separate
attribution head. Five attribution initializations spanning $0.05$--$0.95$
reach exactly the same NLL and remain distinct because likelihood contains no
gradient in that direction. Paired labels recover the correct fraction.

\section{ClosurePairs}

\subsection{Independent nested repeats}

For each fixed $(z,a)$, sample $M$ compatible microstates $X_i$. From every
$X_i$, draw $K$ independent disturbances and obtain $Y_{ik}$. Let $\bar Y_i$
be a row mean and $s_i^2$ its unbiased within-row variance. We estimate
\begin{align}
\widehat \Vproc &= M^{-1}\sum_i s_i^2, \\
\widehat \Valias
  &= S^2(\bar Y_1,\ldots,\bar Y_M)-\widehat\Vproc/K.
\label{eq:nested}
\end{align}

\begin{theorem}[Nested identification]
Before optional non-negativity clipping, both estimators in
Eq.~\ref{eq:nested} are unbiased under conditional exchangeability and finite
second moments.
\end{theorem}

The $1/K$ correction is important: variation among finite-replicate row means
contains residual process variance. Consistency follows as both $M,K$ grow.

\subsection{Crossed reusable disturbances}

Simulators often expose random seeds. Draw $X_1,\ldots,X_M$ and
$E_1,\ldots,E_K$, then evaluate every crossing $Y_{ij}=F(X_i,a,E_j)$. Common
random numbers can reduce Monte Carlo error \citep{glasserman2004monte}, but a
naive comparison of row means is biased in nonlinear dynamics. Write the
orthogonal functional-ANOVA decomposition
\begin{equation}
F(X,E)=m+f_X(X)+f_E(E)+f_{XE}(X,E),
\end{equation}
with component variances $V_X,V_E,V_{XE}$. Then
$\Valias=V_X$ and $\Vproc=V_E+V_{XE}$. If $MS_X,MS_E,MS_{XE}$ are the standard
balanced two-way mean squares, define
\begin{equation}
\widehat V_X=\frac{MS_X-MS_{XE}}{K},\quad
\widehat V_E=\frac{MS_E-MS_{XE}}{M},\quad
\widehat V_{XE}=MS_{XE}. \label{eq:crossed}
\end{equation}

\begin{theorem}[Crossed identification]
The estimators in Eq.~\ref{eq:crossed} are unbiased for their functional-ANOVA
components. Hence $\widehat\Valias=\widehat V_X$ and
$\widehat\Vproc=\widehat V_E+\widehat V_{XE}$ are unbiased.
\end{theorem}

The result follows from $\E MS_X=V_{XE}+KV_X$,
$\E MS_E=V_{XE}+MV_E$, and $\E MS_{XE}=V_{XE}$, the classical balanced
random-effects equations \citep{searle1992variance}. We average raw estimates
before clipping at zero; local training labels are clipped only to keep variance
targets valid.

\begin{corollary}[Finite-budget allocation]
In the balanced Gaussian random-effects model with independent state, noise,
and interaction terms, the unclipped estimators obey
\begin{align}
\Var(\widehat V_X)
 &=\frac{2(V_{XE}+KV_X)^2}{K^2(M-1)}
   +\frac{2V_{XE}^2}{K^2(M-1)(K-1)},\\
\Var(\widehat V_E)
 &=\frac{2(V_{XE}+MV_E)^2}{M^2(K-1)}
   +\frac{2V_{XE}^2}{M^2(M-1)(K-1)},\\
\Var(\widehat V_{XE})
 &=\frac{2V_{XE}^2}{(M-1)(K-1)}.
\label{eq:finite-budget}
\end{align}
For a fixed rollout budget $B=MK$, a pilot plug-in design can therefore
enumerate factor pairs of $B$ and minimize the sum of these variances.
\end{corollary}

This is a classical finite-sample consequence of orthogonal Gaussian mean
squares, not a new REML result \citep{patterson1971recovery,searle1992variance}.
It makes a practical point for world-model evaluation: a square cross is not
universally optimal; allocation should depend on anticipated component sizes
and the target loss. For a downstream finite allocation menu,
Theorem~\ref{thm:allocation-regret} gives a plug-in regret bound and an exact
selection condition in terms of component-estimation error.

\subsection{Nested observation resolutions}

Let $\mathcal G_0\subset\mathcal G_1\subset\cdots\subset\sigma(X)$ be nested
observation sigma-fields and $\mu(X)=\E[Y\mid X,A]$. Define average residual
aliasing $R_r=\E[\Var(\mu(X)\mid\mathcal G_r,A)]$.

\begin{theorem}[Monotone reducibility]
$R_{r+1}\le R_r$, and the removed uncertainty is
\begin{equation}
R_r-R_{r+1}=\E\!\left[\Var\!\left(
\E[\mu(X)\mid\mathcal G_{r+1},A]\mid\mathcal G_r,A
\right)\right]\ge0. \label{eq:monotone}
\end{equation}
\end{theorem}

Thus a resolution curve has a non-increasing aliasing component and, for a fixed
microstate population, a resolution-invariant process floor. This is a closure
diagnostic, not a requirement that a coarse Markov state obey the fine dynamics.

\section{Learning and decision evaluation}

ClosurePairs can score a frozen model or supply auxiliary training targets. Our
Gaussian MLP and mixture-density network (MDN) \citep{bishop1994mdn} contain two
parameter-disjoint branches. The predictive branch minimizes marginal NLL and
outputs a mean/distribution and total variance $\widehat V$. The attribution
branch outputs $\hat\rho(z,a)\in[0,1]$ and is trained only on paired labels:
\begin{equation}
\widehat\Valias=\hat\rho\widehat V,\qquad
\widehat\Vproc=(1-\hat\rho)\widehat V.
\end{equation}
Parameter separation guarantees that paired attribution cannot improve NLL by
construction; this isolates information from predictive capacity. We also test
a stronger shared-representation setting: a CNN encodes pixels, an
action-conditioned GRU rolls out a latent state, and a five-component MDN emits
the future. Lightweight attribution probes read that same recurrent state after
the predictive network is frozen. The exact NLL control is therefore retained
while testing whether paired labels make the mechanism decodable from a
predictive representation.

We report NLL, RMSE, total/component variance error, alias-fraction MAE, and a
sensing decision. If a refined sensor removes aliasing variance $\Delta$ at cost
$c$ under squared-error utility, the Bayes action is ``refine'' iff $\Delta>c$.
The regret of the opposite action is $|\Delta-c|$.

\begin{corollary}[Decision significance]
The optimal refinement action is not identified from the observed kernel in
Proposition~\ref{prop:nonid}. Plugging consistent ClosurePairs estimates into
the rule is decision-consistent whenever $|\Delta-c|$ is bounded away from zero.
\end{corollary}

Our primary deployment target is \emph{not} a lower asymptotic forecast error.
For a context-only policy $\pi(h)=(M,K)$, proper forecast score $S$, fixed
reference sampler $\pi_0$, and predeclared non-inferiority margin $\delta$, the
evaluation problem is
\begin{equation}
\min_{\pi\in\Pi}\;\E[C_\pi(H)]
\quad\text{s.t.}\quad
\E[S(\widehat P_\pi,Y)-S(\widehat P_{\pi_0},Y)]\le\delta .
\label{eq:compute-noninferiority}
\end{equation}
The auditable model-equivalent cost is
$C_\pi=C_0+C_{\rm route}+Mc_X+MKc_E$, where $C_0$ is common encoding,
$c_X$ is the incremental cost of a state particle, and $c_E$ is the cost of a
process branch. A positive deployment claim requires the confidence interval
for the score difference to satisfy non-inferiority and the confidence interval
for $1-\E[C_\pi]/\E[C_{\pi_0}]$ to be positive. Wall-clock, memory, and energy
are implementation-dependent secondary realizations of this same comparison.
Offline ClosurePairs collection is not free: if its one-time cost is
$C_{\rm label}$, the router breaks even only after
$n>C_{\rm label}/(C_{\pi_0}-C_\pi)$ deployment queries, with all terms measured
in the same units.

The amount and the \emph{direction} of computation are distinct. Consider the
additive hierarchy
$Y=\mu+\sqrt{V_{\rm alias}}U+\sqrt{V_{\rm proc}}E$ with standardized,
independent $U,E$. If an ensemble mean uses $M$ state particles and $K$
independent disturbances per state, then
\begin{equation}
R(M,K)=\E[(\widehat\mu_{M,K}-\mu)^2]
=\frac{V_{\rm alias}}{M}+\frac{V_{\rm proc}}{MK}
=V\left(\frac{\rho}{M}+\frac{1-\rho}{MK}\right),
\label{eq:quantity-direction}
\end{equation}
where $V=V_{\rm alias}+V_{\rm proc}$ and
$\rho=V_{\rm alias}/V$. This identity follows by independence of the sampled
state and noise means. Multiplying $V$ changes the minimum cost required to
reach an absolute quality threshold but, under a fixed configuration menu and
cost cap, does not change the risk-minimizing direction. Changing $\rho$ can.
Thus, under finite hierarchical sampling, forecast difficulty governs the useful
compute scale, while the alias/process composition provides complementary
information about its direction. Neither signal alone is a complete allocation
rule.

\begin{figure*}[t]
  \centering
  \includegraphics[width=\textwidth]{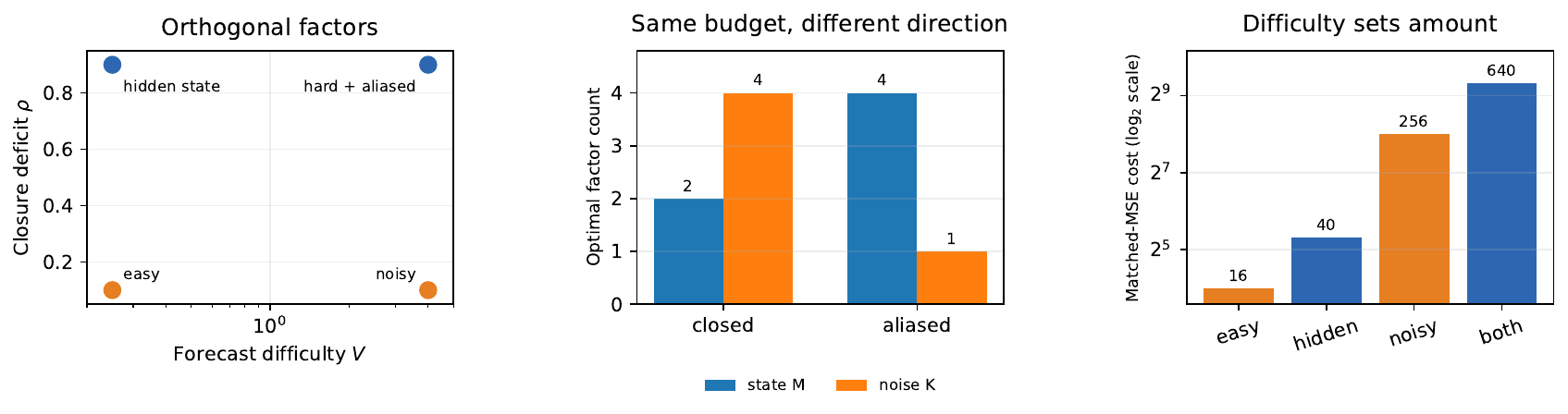}
  \caption{\textbf{Forecast difficulty and dynamical closure are orthogonal
  allocation factors in the controlled hierarchy.} Exact Gaussian systems cross marginal variance
  $V\in\{0.25,4\}$ with alias fraction $\rho\in\{0.1,0.9\}$. Left: all four
  quadrants. Middle: at cost cap 20, equal marginal distributions require
  different state/noise allocations. Right: at matched absolute mean-MSE,
  $V$ changes total cost while $\rho$ changes its composition. Monte Carlo
  verification with $10^5$ trials per cell/configuration agrees with
  Eq.~\ref{eq:quantity-direction} to within $1.1\%$.}
  \label{fig:closure-error-factorial}
\end{figure*}

For probabilistic prediction, the corresponding decision is whether to spend
the next unit of computation on a more informative observation (\textsc{Refine})
or on multiple futures (\textsc{Branch}). ClosurePairs supplies a mechanism
label for this choice. The controlled routing experiment below gives both
routers the same paired rollout matrix and the same predictive heads; it
therefore tests whether variance \emph{composition}, rather than magnitude,
selects the appropriate head. It does not count paired rollouts as free
computation or claim an end-to-end speedup. The pixel experiments instead test
distillation into frozen visual features that do not require paired rollouts at
evaluation time, including a downstream conditional-forecast test.

\section{Experiments}

\subsection{Questions and protocol}

Every experiment tests one of three claims. \textbf{Q1: Identifiability.} Can
systems with the same forecast law have different physical sources of
branching, and do paired interventions recover them at unchanged predictive
score? \textbf{Q2: Transfer.} Does the identified split remain decodable from
ordinary test observations and frozen visual context without paired futures at
test time? \textbf{Q3: Decision value.} Does source composition determine
state-versus-noise allocation when forecast difficulty does not, and how does
it compare with standard supervision? Learned experiments use held-out
intervention groups. Confidence intervals bootstrap
independent seed-level differences; Wilcoxon tests are two-sided and paired.
Appendix~\ref{app:implementation} gives all hyperparameters.

\subsection{Exact and learned Gaussian systems}

Equation~\ref{eq:family} uses $V=0.64$ and
$\lambda\in\{0.1,0.3,0.5,0.7,0.9\}$. The observed oracle is identical across
the family. The learned Gaussian MLP reduces alias-fraction MAE from $0.2400$
for the arbitrary observational $0.5$ split to $0.0150$ with paired supervision
($15.96\times$), with mean test-NLL difference $0.000000000$. Component-level
alias-variance MAE, which also includes noisy local variance prediction, falls
from $0.1755$ to $0.0874$.

This comparison does not claim that $0.5$ is the strongest possible guess for
every member. Proposition~\ref{prop:nonid} says no observational rule can be
correct uniformly: changing its inductive bias merely selects another point on
the same likelihood ridge.

\subsection{Equal-budget estimation and deployment}

We compare three data-collection protocols at exactly $B=MK$ simulator calls
per context: crossed ClosurePairs with the two-way moment/REML equations,
independent nested repeats with the one-way finite-repeat correction, and naive
common random numbers (CRN), which compares crossed row means but does not
estimate interaction. The nonlinear simulator is
\[
Y=\sqrt{V_X(z)}X+0.15(X^2-1)+\sqrt{V_E(z)}E
  +\sqrt{0.2}\,XE ,
\]
where coefficients are normalized so $V_X(z)+V_{\mathrm{proc}}(z)=1$ for
every context and $V_{XE}=0.2$. Thus total variance has no routing signal, and
the interaction cannot be ignored. For each budget we report the best
allocation available to each method, preventing an arbitrary $M/K$ choice from
favoring ClosurePairs.

\begin{figure*}[t]
  \centering
  \includegraphics[width=\textwidth]{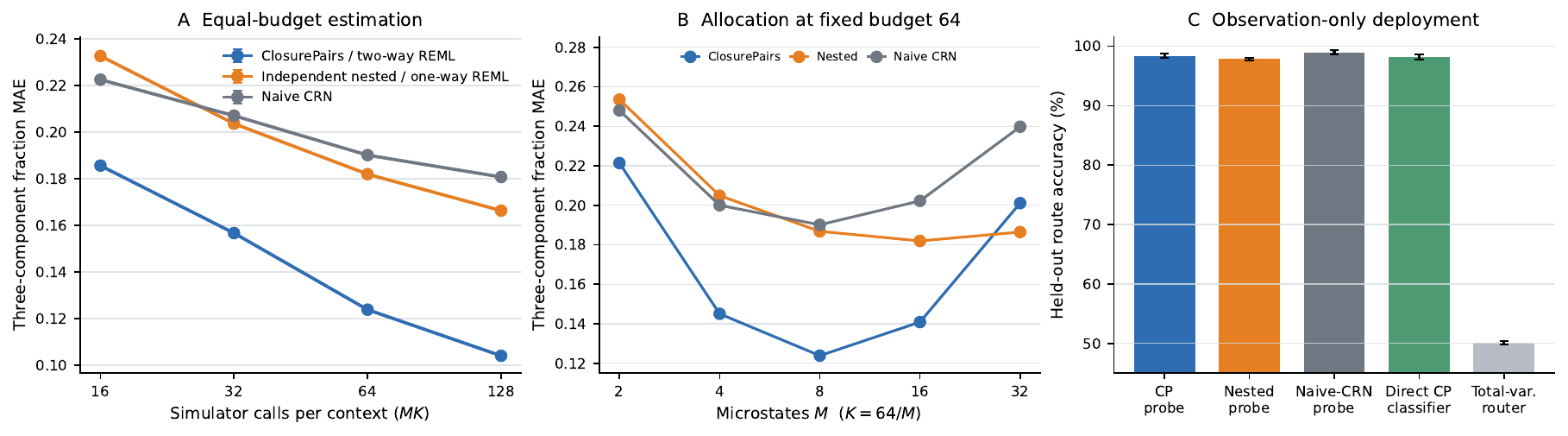}
  \caption{\textbf{Equal-budget protocol comparison and deployment.}
  (a) Best three-component source-fraction MAE at each common rollout budget;
  bands are 95\% intervals over 80 trials with 256 contexts each.
  (b) Allocation sensitivity at $B=64$.
  (c) Held-out routing after offline label generation; test inputs contain only
  the current observation. Bars are means over 30 independent splits.}
  \label{fig:fixed-budget}
\end{figure*}

Figure~\ref{fig:fixed-budget}a shows that ClosurePairs has the lowest
three-component fraction MAE at every tested budget; at $B=64$ it is $0.124$
versus $0.182$ for independent nesting and $0.190$ for naive CRN. Panel b
confirms that extreme factor ratios are inefficient and that interaction cannot
be ignored.

We next use intervention rollouts only to create \emph{offline} training labels.
Each of 30 splits has 384 training contexts and a $16\times16$ training cross
per context. At test time every probe receives only the current scalar
observation $z$---no reset, seed replay, paired statistic, or future rollout.
The ClosurePairs regression probe reaches $98.42\%$ route accuracy, compared
with $97.84\%$ for equal-budget nested supervision, while a total-variance
router remains at $50.09\%$. Naive CRN reaches $99.00\%$ on this particular
binary boundary despite biased source attribution. Thus ClosurePairs is most
useful when the target is the complete, reusable decomposition; it is not
uniformly optimal for every finite-budget classifier.

\subsection{Exact-marginal MetaWorld compute allocation}
\label{sec:metaworld-compute}

We next test the compute interpretation directly in public MetaWorld Push and
Push-Wall. For each nominal state, hidden and future object-velocity impulses
enter the same MuJoCo channel with scales $(0.8,0.4)$ or $(0.4,0.8)$. Reusing one centered
four-point design makes the two crossed tables transposes: their empirical
future distributions match although their alias fractions differ by
$0.559/0.533/0.511$ on three 64-pair Push tests and by $0.545$ on a 64-pair
Push-Wall test. The visible condition is temporal
sensor resolution. A Closure-supervised probe on frozen JEPA-WM context
features routes $99.2/99.2/89.8\%$ on two ID and one camera-plus-action OOD
Push test and $100\%$ on Push-Wall, while a generous output-only probe remains
at chance. Thus matched forecast distributions do not reveal the mechanism
direction, while intervention supervision makes it decodable from the current
visual context.

With a learned hierarchical stochastic readout, fixed-reference CRPS and
Closure are nearly orthogonal on Push ID/OOD (Spearman $-0.058/-0.376$).
Accordingly, a marginal-output allocator has $50\%$ mechanism-direction
accuracy and resolves no state-versus-noise twin pair, whereas Closure
supervision reaches $79.7/75.8\%$. This is the central empirical distinction:
forecast difficulty alone misses
compute direction, whereas Closure supplies the missing mechanism target.

\begin{figure*}[t]
  \centering
  \includegraphics[width=0.82\textwidth]{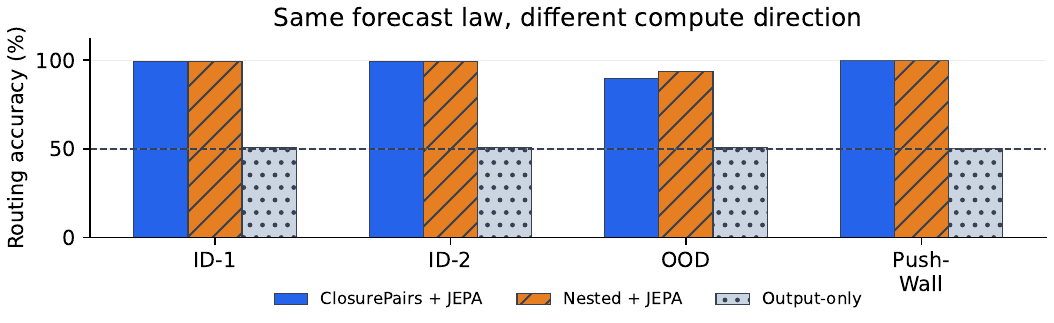}
  \caption{\textbf{Same forecast distribution, different compute direction.}
  Each split has 64 exact-marginal physical twin pairs; Push-Wall is a distinct
  second task. Test-time probes receive only the current frozen JEPA-WM context
  features. The dashed line is chance.}
  \label{fig:metaworld-routing}
\end{figure*}

\subsection{Independent ManiSkill stochastic-RSSM confirmation}
\label{sec:maniskill-confirmation}

We independently instantiate the intervention contract in public ManiSkill
PushCube. Within each pair, both twins use the same object, physics, reset seed,
zero-action rollout, and pooled initial-velocity support. A coarse observation
operator leaves large compatible-state velocity variation and a fine operator
leaves little; the fresh velocity disturbance receives the complementary
coefficient. With the coefficients swapped, the two $8\times8$ crossed tables
are transposes.
Full-state replay after an intervening robot rollout is accurate to
$4.55\times10^{-13}$ in maximum absolute state error. On held-out contexts the
crossed alias-fraction gap is $0.707$, while the between-twin sliced-Wasserstein
distance is numerically zero.

The deployable observation is only a $32\times32$ RGB image. Coarse images are
sampled at $8\times8$ before reconstruction; no sensor resolution, physical
parameter, or mechanism metadata is provided. A physical-camera OOD split
changes the actual cube half-size and hence contact dynamics, camera angle and
scale, and palette. We then fit an image-conditioned stochastic RSSM and freeze
it. Offline probes receive ClosurePairs labels, but test-time probes receive
only predicted samples, the RSSM latent, or RGB; none receives paired futures.
Table~\ref{tab:maniskill-rssm} reports five predeclared training seeds. Forecast
quality is practically identical: the coarse--fine energy-score gap is
$5.6\times10^{-6}$ ID and $-1.9\times10^{-6}$ OOD. Output-only and latent
routing remain at chance, whereas the RGB Closure probe selects the equal-call
$(M,K)\in\{(4,1),(1,4)\}$ direction in every run and removes the positive
finite-sampling regret of either fixed direction. The direct allocation
baseline also reaches $100\%$ and zero regret, while a neutral-sensor control is
exactly $50\%$. This confirmation therefore supports metadata-free RGB routing
and physical-parameter transfer, with parity to a task-specific allocator; it
does not establish universal allocation superiority. Direct supervision is an
optimistic, decision-specific baseline. We test reuse over five unseen
equal-call menus and setup-cost regimes. Closure predicts the three crossed
fractions once and evaluates finite-population state/noise/interaction risk for
each new menu, without new oracle labels. Table~\ref{tab:maniskill-transfer}
shows that this reduces downstream re-supervision rather than proving fewer
simulator calls for initial label collection. The deterministic
state-to-RGB operator is reproducible but is not ManiSkill's native camera
renderer.

\begin{table}[t]
  \centering
  \small
  \caption{\textbf{ManiSkill PushCube confirmation over five RSSM seeds.}
  Intervals bootstrap seed-level means. Regret is excess empirical energy score
  relative to the per-context oracle allocation.}
  \label{tab:maniskill-rssm}
  \begin{tabular}{lcccc}
    \toprule
    & \multicolumn{2}{c}{ID} & \multicolumn{2}{c}{Physical-camera OOD} \\
    \cmidrule(lr){2-3}\cmidrule(lr){4-5}
    Test-time signal & Acc. (\%) & Regret & Acc. (\%) & Regret \\
    \midrule
    Predicted marginal output & $50.4$ & $3.84\times10^{-4}$ & $44.2$ & $4.43\times10^{-4}$ \\
    Frozen RSSM latent + Closure labels & $51.7$ & $3.71\times10^{-4}$ & $47.9$ & $4.15\times10^{-4}$ \\
    RGB + Closure labels & $100.0$ & $0$ & $100.0$ & $0$ \\
    RGB + direct allocation labels & $100.0$ & $0$ & $100.0$ & $0$ \\
    Neutral-sensor RGB control & $50.0$ & --- & $50.0$ & --- \\
    \bottomrule
  \end{tabular}
\end{table}

\begin{table}[t]
  \centering
  \small
  \caption{\textbf{Zero-shot supervision transfer over five unseen allocation
  menus.} Values average the fixed menu set. Retraining uses 96 new oracle
  labels per menu; Closure uses none.}
  \label{tab:maniskill-transfer}
  \begin{tabular}{lccccc}
    \toprule
    & New labels & \multicolumn{2}{c}{ID} & \multicolumn{2}{c}{Physical-camera OOD} \\
    \cmidrule(lr){3-4}\cmidrule(lr){5-6}
    Method & & Acc. (\%) & Regret & Acc. (\%) & Regret \\
    \midrule
    Closure zero-shot & $0$ & $92.5$ & $0.00070$ & $90.4$ & $0.00105$ \\
    Direct frozen & $0$ & $37.9$ & $0.02590$ & $32.9$ & $0.02600$ \\
    Direct retrained & $480$ & $97.1$ & $0.00014$ & $92.1$ & $0.00052$ \\
    \bottomrule
  \end{tabular}
\end{table}

The Closure--frozen-direct regret difference is $-0.0252$ ID
(scenario-bootstrap 95\% interval $[-0.0375,-0.0100]$) and $-0.0249$ OOD
($[-0.0379,-0.00981]$), corresponding to $97.3\%$ and $95.9\%$ relative regret
reductions. Retrained direct is a slightly stronger task-specific upper bound,
as expected, but requires 480 new labels.

\FloatBarrier
\section{Related work}

\paragraph{Stochastic world-model evaluation.}
STORI separates stochastic perturbation types and finds failures of model-based
RL agents \citep{barsainyan2025stori} and states the same total-variance split;
that identity is not our contribution. Infoprop separates epistemic and
aleatoric uncertainty to decide when a model rollout should stop
\citep{frauenknecht2025rollouts}; ClosurePairs instead asks whether finite
compute should resolve the current state or branch over process noise. CaliBench evaluates physical outcome
distributions \citep{anonymous2026calibench}, while recent work cautions against
an intrinsic aleatoric--epistemic dichotomy
\citep{bickfordsmith2024rethinking}. ClosurePairs instead adds a reset protocol
with microstate and repeated-noise interventions, defines components relative
to an explicit state boundary, and estimates their interaction.

\paragraph{Stochastic simulators and experimental design.}
Replication and variance-component design are mature
\citep{baker2022analyzing}: stochastic kriging models simulation noise
\citep{ankenman2010stochastic}, CRN couples runs
\citep{schruben2011common,glasserman2004monte}, REML estimates components
\citep{patterson1971recovery,searle1992variance}, and Sobol decompositions apply
to stochastic simulators \citep{sobol1993sensitivity,lemaitre2015variance}.
ClosurePairs claims these tools only as ingredients; its contribution is a
world-model intervention contract, equal-budget comparison, and distillation
into a test-time observation-only rule.

\paragraph{Physical inference and paired interventions.}
Physion++, POKEWORLD, and What-If World test hidden physical parameters or
semantic interventions \citep{tung2023physion,tan2026latent,cai2026whatif}.
ClosurePairs instead intervenes on the microstate fiber and disturbance to
identify a reset-based variance estimand.

\paragraph{Closure and memory.}
Coarse dynamics motivate memory and latent-state models
\citep{mori1965transport,zwanzig1961memory,sanderse2024closure,ruiz2025memory,ilersich2025stochastic};
we evaluate their declared state boundary rather than propose another model.

\section{Limitations and conclusion}

The main conclusion is simple: a predictive distribution can identify how hard
the future is to forecast without identifying why it branches. ClosurePairs
supplies that missing mechanism information. Under finite hierarchical sampling,
forecast difficulty governs the useful compute scale, while the alias/process
composition provides complementary information about its direction. This
distinction survives equal likelihood, nonlinear interaction, exact-marginal
MetaWorld twins, an independent ManiSkill stochastic-RSSM confirmation, and
zero-shot reuse across changed hierarchical-compute menus.

The result is an identification and evaluation claim, not universal routing
superiority. ClosurePairs requires reset access, repeated disturbances, and a
declared state boundary. Evidence remains simulator-based and costs are
model-equivalent. We do not claim a new predictive architecture or universal
compute saving. ClosurePairs is useful when the scientific target is the
reusable cause of branching rather than one fixed forecast score.

\bibliography{references}
\bibliographystyle{arxiv_style}

\subsection*{AI use statement}
A large language model assisted literature triage, code and figure scaffolding,
and language editing. The author checked all outputs and takes responsibility.

\subsection*{Ethics and reproducibility}
The work uses no human data or safety-critical deployment. The supplement and
code include proofs, generators, hyperparameters, seeds, tests, and scripts.

\appendix
\section{Proofs}
\label{app:proofs}

\subsection{Proof of Proposition~\ref{prop:nonid}}

Conditioned on $(Z,A)$, the two random terms in Eq.~\ref{eq:family} are
independent zero-mean Gaussians with variances $\lambda V$ and
$(1-\lambda)V$. Their sum is $\mathcal N(0,V)$ for every $\lambda$. Conditioning
further on the full state, which contains $U$, leaves variance $(1-\lambda)V$;
the variance of the conditional mean over compatible $U$ is $\lambda V$.

Suppose an observational estimator $T_n$ were consistent for $\Valias$ at all
$\lambda$. Its sampling distribution is identical for all $\lambda$, because
the joint observed law of $(Z,A,Y)$ is identical. It therefore cannot converge
in probability to both $\lambda_1V$ and $\lambda_2V$ for
$\lambda_1\ne\lambda_2$, a contradiction. The same argument applies to
$\Vproc$ and to any statistic or learned representation measurable with respect
to the observational sample.

\subsection{Proof of nested identification}

Let $\mu_i=\E[Y_{ik}\mid X_i]$ and
$\sigma_i^2=\Var(Y_{ik}\mid X_i)$. Unbiased within-row sample variance gives
$\E[s_i^2\mid X_i]=\sigma_i^2$, so
$\E\widehat\Vproc=\E_X\sigma_X^2=\Vproc$. Conditional independence within a row
gives
\[
\Var(\bar Y_i)=\Var_X(\mu_X)+\E_X[\sigma_X^2]/K
=\Valias+\Vproc/K.
\]
The sample variance of i.i.d. row means is unbiased for this quantity. Subtracting
$\widehat\Vproc/K$ proves the claim. Optional clipping creates a small boundary
bias but enforces a valid finite-sample variance.

\subsection{Proof of crossed identification}

Define
\begin{align*}
m&=\E F(X,E),\\
f_X(x)&=\E_EF(x,E)-m,\\
f_E(e)&=\E_XF(X,e)-m,\\
f_{XE}(x,e)&=F(x,e)-m-f_X(x)-f_E(e).
\end{align*}
Every nonconstant component has mean zero, and $f_{XE}$ has zero conditional mean
given either argument. Hence the three components are mutually orthogonal.
Moreover,
\[
\Valias=\Var(f_X)=V_X,
\quad
\Vproc=\E_X\Var_E(f_E+f_{XE}\mid X)=V_E+V_{XE}.
\]
Expanding the balanced row, column, and residual sums of squares and taking
expectations gives
\[
\E MS_X=KV_X+V_{XE},\quad
\E MS_E=MV_E+V_{XE},\quad
\E MS_{XE}=V_{XE}.
\]
Solving these three equations yields Eq.~\ref{eq:crossed}.

\subsection{Finite-budget calculation}

Under the Gaussian random-effects assumptions, the row, column, and interaction
mean squares are independent scaled chi-squared variables:
\[
\frac{(M-1)MS_X}{V_{XE}+KV_X}\sim\chi^2_{M-1},\quad
\frac{(K-1)MS_E}{V_{XE}+MV_E}\sim\chi^2_{K-1},
\]
and
\[
\frac{(M-1)(K-1)MS_{XE}}{V_{XE}}
\sim\chi^2_{(M-1)(K-1)}.
\]
Their variances are twice the squared scales divided by their degrees of
freedom. Substitution into Eq.~\ref{eq:crossed}, using independence of the
orthogonal quadratic forms, gives Eq.~\ref{eq:finite-budget}. These are
unclipped componentwise variances; clipping trades variance for boundary bias.
For fixed $B=MK$, evaluating their sum over integer factor pairs gives the
stated pilot plug-in allocation.

\begin{theorem}[Plug-in allocation regret]
\label{thm:allocation-regret}
Let $\mathcal C$ be a finite cost-feasible menu of $(M,K)$ allocations. With
fresh independent process branches per state particle, write
$R_{M,K}=\Valias/M+\Vproc/(MK)$. If
$\max(|\widehat\Valias-\Valias|,
|\widehat\Vproc-\Vproc|)\le\epsilon$ and the plug-in rule minimizes
$\widehat R$ over $\mathcal C$, then
\[
R_{\widehat M,\widehat K}-\min_{(M,K)\in\mathcal C}R_{M,K}
\le 2\epsilon L,\qquad
L=\max_{(M,K)\in\mathcal C}\left(\frac1M+\frac1{MK}\right).
\]
If the true risk gap between the best and second-best allocations exceeds
$2\epsilon L$, the plug-in rule selects the true optimum.
\end{theorem}

For every allocation, $|\widehat R_{M,K}-R_{M,K}|\le\epsilon L$.
Let $(M^*,K^*)$ minimize true risk. Optimality under the estimated components
gives
$R_{\widehat M,\widehat K}\le
\widehat R_{\widehat M,\widehat K}+\epsilon L\le
\widehat R_{M^*,K^*}+\epsilon L\le R_{M^*,K^*}+2\epsilon L$.
The margin statement follows because no suboptimal allocation can reverse its
ordering when both estimated risks move by at most $\epsilon L$.

\subsection{Proof of monotone reducibility}

Apply conditional total variance to $\mu(X)$ under nested sigma-fields:
\begin{align*}
\Var(\mu\mid\mathcal G_r,A)
&=\E[\Var(\mu\mid\mathcal G_{r+1},A)\mid\mathcal G_r,A]\\
&\quad+\Var(\E[\mu\mid\mathcal G_{r+1},A]\mid\mathcal G_r,A).
\end{align*}
Taking expectations proves Eq.~\ref{eq:monotone}. Non-negativity of conditional
variance gives monotonicity.

\section{Implementation details}
\label{app:implementation}

\paragraph{Analytic linear system.}
The full state is $X=(Z,U_1,U_2,U_3)$ and observations progressively reveal the
first $r$ hidden coordinates. The future is
$Y=0.9Z+\beta^\top U+0.6A+0.45E$, with
$\beta=(1.2,-0.7,0.35)$ and hidden standard deviations $(0.8,0.6,0.5)$.
The analytic aliasing curve is $(1.128625,0.207025,0.030625,0)$ and the process
floor is $0.2025$. Estimated values are
$(1.127858,0.204056,0.031285,0)$ and approximately $0.199$--$0.204$. In this
additively separable system, crossed common random numbers reduce alias-estimator
RMSE by $2.72$--$2.98\times$ for 16--128 groups.

\paragraph{Equal-budget benchmark.}
The contextual simulator uses standard-normal $X,E$, an $X^2$ nonlinearity
with coefficient $0.15$, and interaction variance $0.2$. Context-dependent
main-effect coefficients are normalized analytically so that total variance is
one. For budgets $16,32,64,128$, we enumerate every integer $(M,K)$ with
$M,K\ge2$. Each method receives exactly $MK$ calls. Reported intervals aggregate
80 independent trials with 256 contexts. Independent nested repeats use fresh
disturbances in every row; crossed methods reuse the same $K$ disturbances.
Balanced two-way Gaussian REML and ANOVA have the same unconstrained moment
equations here; estimates are clipped only after solving them. Methods that do
not identify interaction report it as zero for the prespecified
three-component loss rather than receiving oracle interaction labels.

For deployment, each of 30 seeds uses 384 training and 4,096 test contexts.
Only training contexts receive $16\times16$ rollouts. Cubic ridge probes predict
the continuous state fraction from scalar $z$; the direct baseline instead fits
logistic regression to ClosurePairs route labels. Every test method receives
only $z$. Paired $t$ and Wilcoxon tests operate on the 30 seed-level metrics.

\paragraph{Learned Gaussian experiment.}
Each system uses 192 train groups with 16 microstates and eight disturbances,
and 256 test groups with 20 microstates and ten disturbances. The MLP has two
64-unit SiLU layers (48 units in the reported run), Adam learning rate
$3\times10^{-3}$, 300 full-batch epochs, and seeds 11, 23, 37. The paired loss is
mean-squared error on the ANOVA alias fraction. The likelihood and attribution
networks do not share parameters.

\paragraph{Langevin MDN.}
We use $dt=0.02$, friction $0.5$, thermal scale $0.7$, initial velocity standard
deviation $0.8$, and action scale $0.25$. Every condition uses 256 train groups,
16 microstates, eight disturbances, and 450 epochs. The five-component MDN has
two 64-unit SiLU layers, Adam learning rate $2\times10^{-3}$, and seeds 13, 29,
43. ID tests use 320 groups, 20 microstates, and ten disturbances. OOD tests draw
640 groups with action scale $0.55$ and retain only $|a|>0.30$. The sensing cost
is $0.05$.

\paragraph{Pendulum.}
We follow Pendulum-v1 constants $g=10$, $m=l=1$, $dt=0.05$, maximum torque 2,
and maximum angular speed 8, adding Gaussian torque noise with standard deviation
$0.35$. Each of ten seeds uses 192 groups, 24 microstates, and 16 crossed noise
sequences at horizons 10, 25, 50, 100.

\paragraph{Pixel recurrent model.}
Images are $32\times32$ deterministic grayscale renders. At each horizon
$10,25,100$, training uses 128 angle-only groups with 12 compatible microstates
and eight crossed noise sequences; the velocity-refined predictor uses 1024
groups with three microstates. ID and OOD tests each use 128 groups per horizon,
16 microstates, and ten disturbances. The encoder has three convolutional
blocks, the GRU state has 64 units and eight action/horizon-conditioned rollout
steps, and the emission is a five-component Gaussian mixture. We train three
ensemble members for 120 epochs, then freeze the first member and train its two
shared-state attribution probes for 220 epochs. Monte Carlo dropout uses the
same base predictor. Seeds are 11, 23, 37, 53, 71, 83, 97, 109, 127, and 149.
Active-sensing thresholds are selected only on held-out contexts; sensor cost is
$0.065$.

For clarity, the ensemble baseline maps the across-member variance of predictive
means to the aliasing proxy and the mean within-member MDN variance to the
process proxy; its reported alias fraction is their ratio. MC dropout applies
the identical mapping across 24 stochastic forward passes. These are
conventional epistemic/aleatoric proxies, not identifiable estimators of our
physical components.

\paragraph{Learned neural-SDE sensor intervention.}
For each of ten seeds, the attribution set contains 160 contexts and the test
set 120 disjoint contexts. Context features contain two physical-condition
coordinates, their squares, position, action, and three nuisance coordinates;
the true split is a fixed nonlinear function of the first two coordinates.
Every training label uses eight compatible initial velocities and eight thermal
impulses. The independent-nested baseline uses a separately generated equal-cost
table. The observational system-identification baseline instead receives eight
ordinary trajectories of eight transitions, fits the split by Gaussian
quasi-likelihood over 101 candidates using the locally linearized double-well
covariance, and distills that estimate through the same adapter. The one-step
variance is $4\times10^{-5}$, $dt=0.02$, $\gamma=0.5$, actions are uniform on
$[-0.25,0.25]$, initial positions on $[-0.4,0.4]$, and the forecast horizon is
80. The drift MLP has two 64-unit SiLU layers and trains for 500 full-batch AdamW
epochs on an independently generated 6,000-transition set; Gaussian noise with
standard deviation $0.15$ is added to acceleration targets. The source adapter
is a 300-tree random forest with minimum leaf size four. Hyperparameters are
fixed for all seeds. Conditional CRPS uses common empirical truth samples across
methods; the reported significance test pairs seed-level mean CRPS.

The RGB extension uses 480 training and 120 test contexts per seed. Each current
frame is $32\times32$ RGB; a previous frame separated by $0.2$ time units is
used only for centroid-based velocity sensing. The frozen visual encoder returns
per-channel moments and quantiles, a $4\times4$ average-pooled grid, and eight
horizontal Fourier magnitudes per channel. Each source estimator receives these
features and action through a 400-tree random forest with minimum leaf size four
and feature fraction $0.7$. Closure and independent nesting each use an
$8\times8$ table per training context; the multi-step baseline retains eight
eight-transition trajectories. All label probes, including the privileged
true-split diagnostic, use the same encoder and regressor. Test-time source
prediction uses one current frame and no simulator reset, disturbance seed, or
future rollout.

For the public-representation replacement, all $39{,}687{,}136$ parameters of
the JEPA-WM MetaWorld epoch-50 checkpoint remain frozen. We bilinearly resize the
current RGB frame to $224\times224$, pool its $16\times16$ DINOv2 patch map to a
$4\times4$ grid, fit a 64-dimensional PCA on training contexts only, append the
declared action, and use the same random-forest hyperparameters. The previous
frame is deliberately excluded from this source probe because it is the sensing
intervention being evaluated. We do not call the JEPA action-conditioned
predictor in this experiment; every method retains the common learned neural
SDE rollout.

The state-feature version removes $84.0\%\pm1.7\%$ of the fixed model's excess
conditional CRPS. Over ten independent RGB seeds, the generic-encoder Closure probe
reaches source-fraction MAE $0.069\pm0.002$, removes $85.0\%\pm1.0\%$ of excess
CRPS, and lowers conditional variance relative error from $0.486\pm0.019$ to
$0.158\pm0.004$ (bootstrap 95\% CRPS-gain CI $[0.00955,0.01158]$; two-sided
Wilcoxon $p=0.00195$). The frozen JEPA representation reaches fraction MAE
$0.082\pm0.002$, removes $76.4\%\pm1.6\%$ of excess CRPS, and gives variance
error $0.208\pm0.008$ (gain CI $[0.00855,0.01040]$; $p=0.00195$). Equal-cost
independent nesting reaches $0.177\pm0.006$ variance error but is statistically
tied with crossing in CRPS ($p=0.432$). The correctly specified multi-step MLE
reaches $0.116\pm0.004$ and outperforms ClosurePairs ($p=0.00195$).

\begin{figure}[h]
  \centering
  \includegraphics[width=\linewidth]{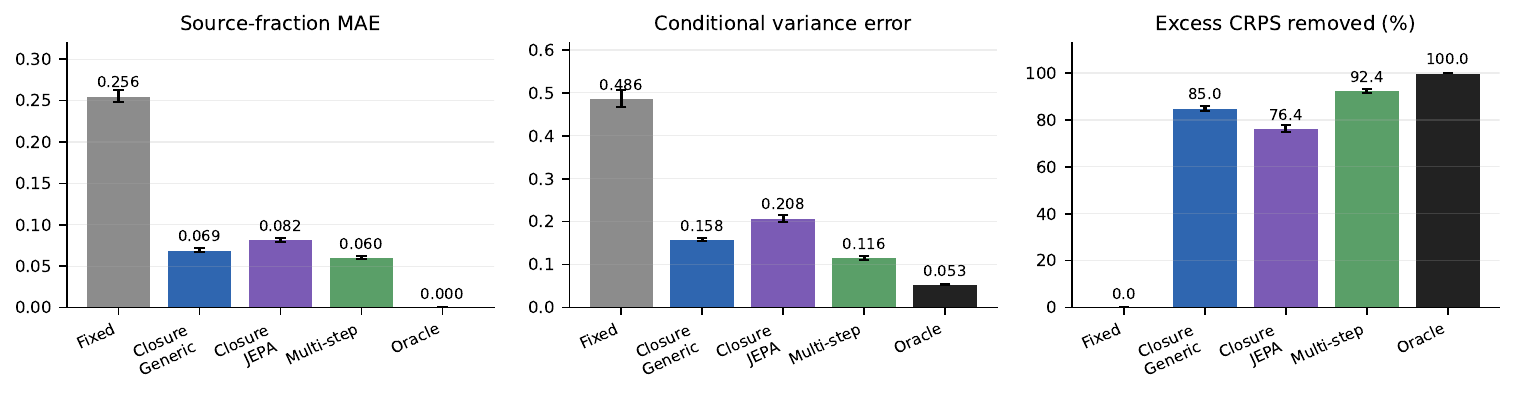}
  \caption{\textbf{RGB-conditioned physical forecast after sensing velocity.}
  Lower is better in the first two panels; higher is better on the right. Bars
  show means and standard errors over ten independent seeds, with values above
  every bar. ``Generic'' is the deterministic visual encoder; ``JEPA'' uses the
  frozen public JEPA-WM representation. Closure supervision removes
  hidden-velocity uncertainty while the learned-drift neural SDE retains thermal
  diffusion.}
  \label{fig:neural-sde-sensor}
\end{figure}

\paragraph{Matched-variance routing.}
Each of seeds 2027--2031 uses 6,000 training and 2,000 test trajectories at a
12-step horizon. ClosurePairs uses four compatible microstates crossed with four
reusable disturbances. The REFINE head is a two-layer Gaussian trajectory MLP
conditioned on four frames; the BRANCH head is a two-layer five-component
mixture trajectory MLP conditioned on the current frame. Both also receive the
same four-dimensional log-variance summary. A 48-unit probe reads that summary,
while the total-variance baseline thresholds only its final coordinate. The
threshold is selected on training data. All three networks train for 1,800
minibatch steps. The animation uses a held-out matched pair, but all reported
values aggregate the complete test sets.

\paragraph{Exact-marginal JEPA-WM compute benchmark.}
The MetaWorld Push benchmark uses 24/16/64 physical twin pairs for
train/validation/test and an independent 64-pair ID replication. A third
64-pair test changes the camera from \texttt{corner2} to \texttt{corner} and
the expert-policy gain range from $[0.8,1.0]$ to $[0.5,0.7]$. Every physical
pair produces two contexts with swapped hidden-state/future-disturbance scales.
The frozen epoch-50 JEPA-WM encoder supplies temporal-statistic features; a
PCA--ridge source probe is fit using train and validation labels, while the
hierarchical stochastic readout is fit only on the 48 training contexts. It
uses ridge mean/log-scale heads and a 32-nearest-neighbor library of normalized
training residuals. Forecasts use 32 fixed random projections and 100 sampling
repetitions per allocation. The candidate set is
$M,K\in\{1,2,4\}$, $c_X=4$, $c_E=1$, and $C=M c_X+MKc_E$.
The budget projection sees only predicted source fraction and costs; it does not
read test outcomes. The fixed allocation is selected on validation, and
confidence intervals bootstrap the 64 independent twin-level differences with
20,000 resamples. For the learned readout, matched-score compute savings at
$c_X/c_E\in\{1,2,4,8,16\}$ are respectively
$16.7/25.4/18.2/11.1/19.8\%$; an oracle-sampler version is retained only as a
feasibility diagnostic. The direct baseline labels each offline context by the
lowest-CRPS feasible allocation, selects a 500-tree random forest with minimum
leaf size in $\{1,2,4,8\}$ on validation, and refits 800 trees on train plus
validation. It therefore uses a decision- and cost-specific label rather than a
reusable physical source fraction.

A stronger risk-surface baseline predicts relative sliced CRPS for all nine
configurations from the same frozen context features. PCA dimension and ridge
penalty minimize validation regret averaged over
$c_X/c_E\in\{1,2,4,8,16\}$; the surface is then reused without retraining at
every cost. Across six tests and five ratios it wins 20 of 30 comparisons and
averages $17.25\%$ matched-quality saving versus ClosurePairs' $15.90\%$.
Thus the physical decomposition is not required for optimal allocation on this
benchmark. Its distinct target is identifiable and reused across attribution,
sensing, scores, and costs. We also average each of the nine score labels over
$R\in\{1,2,4,8,16,32,100\}$ sampling repeats. On ID, the risk surface already
saves $16.9\%$ at $R=1$ (49 forecast draws per context), close to ClosurePairs'
$18.2\%$ from a 16-cell intervention table. Simulator cells and learned-generator
draws have different costs, so we report rather than equate these counts.

The corresponding hardware-independent counts are $3.406/3.438/3.484/3.219$
state particles and $6.000/6.125/5.938/7.031$ terminal branches per context,
versus $4$ and $8$ for matched fixed $(4,2)$ sampling. A randomized nine-block,
1,000-call timing audit of the NumPy hierarchical sampler fits
$t=\beta_0+\beta_XM+\beta_EMK$ with $R^2=0.9996$ and yields
$15.2$--$25.7\%$ lower sampler time across six tests. It excludes common frozen-JEPA
encoding, preparation, routing, and scoring, so it is a directional
implementation check rather than the primary compute definition.

We repeat the complete admission and 100-repeat learned-readout protocol on
MetaWorld Push-Wall with an independent 24/16/64-pair split. Its test mechanism
gap is $0.545$; the forecast-energy difference is $4.1\times10^{-7}$ (95\% CI
$[-9.8\times10^{-7},2.1\times10^{-6}]$), the between/within sliced-Wasserstein
ratio is $1.02\times10^{-4}$, and output-only route accuracy is $50.0\%$.
Closure-aware routing uses $19.906$ units to match cost 24 ($17.1\%$ saving);
under the per-context cap it uses $18.219$ units ($24.1\%$ saving), with a
positive paired CRPS-difference interval $[0.00241,0.00473]$.

\paragraph{ManiSkill RSSM confirmation.}
We use ManiSkill 3.0.1 PushCube with the CPU PhysX backend, zero
robot-initialization noise, and a training/ID cube half-size of $0.02$. The OOD
half-size is $0.03$; its camera angle changes from $[-12,12]$ to $[32,48]$
degrees, with disjoint camera-scale and palette ranges. A deterministic
top-down state-to-RGB operator produces $32\times32$ images; the coarse branch
first samples $8\times8$, and neither model nor probes receive operator
metadata. The input alias fractions are $0.9/0.1$. Each of 48 training, 24 ID,
and 24 OOD contexts per class uses eight compatible velocities crossed with
eight reusable kicks over an eight-step zero-action rollout. The RSSM has a
64-dimensional deterministic state, 12-dimensional
stochastic state, four recurrent steps, and 16 mixture samples; AdamW trains it
for 180 full-batch epochs at learning rate $2\times10^{-3}$. The fixed seeds are
23011--23015. Output and latent baselines use 300-tree minimum-leaf-two
regressors. Closure and direct RGB probes use the same standardized linear
regressor over fixed multiscale spatial-frequency statistics derived solely
from pixels. The former predicts continuous state, noise, and interaction
fractions; the latter predicts the oracle choice for one fixed decision. Allocation scores
compare the equal-call choices $(4,1)$ and
$(1,4)$. Confidence intervals bootstrap the five seed-level means with 20,000
resamples.

The transfer audit freezes both probes after this menu and evaluates five
unseen menus with products 8, 12, or 16 and declared state/noise setup costs.
For a menu $(M,K)$, Closure predicts normalized mean-estimation risk with
$V_s g(M)+V_n g(K)+V_{sn}g(M)g(K)$ plus declared setup cost, where
$g(q)=(8-q)/(7q)$. Empirical oracles average 512 subsampling repeats. The direct
upper bound is retrained on 96 new oracle labels per menu; frozen direct and
Closure receive none. Transfer intervals descriptively bootstrap the five
fixed scenario-level values.

\paragraph{Compute.}
The attribution experiments run on CPU; the matched-variance routing runs use an
NVIDIA RTX 4060 Laptop GPU. The verified environment uses Python 3.10.20,
PyTorch 2.7.0, NumPy 2.2.6, SciPy 1.15.3, scikit-learn 1.7.2, and Matplotlib
3.10.9. The current complete run passes 144 tests; three optional-dependency
tests are skipped. Coverage includes analytic recovery, equal-budget baselines,
boundary controls, and nonlinear interaction recovery. The pixel
recurrent run checkpoints after each seed. Exact commands are in the repository.

\section{Additional numerical results}

\subsection{Nonlinear multi-step attribution}

\begin{figure}[h]
  \centering
  \includegraphics[width=\linewidth]{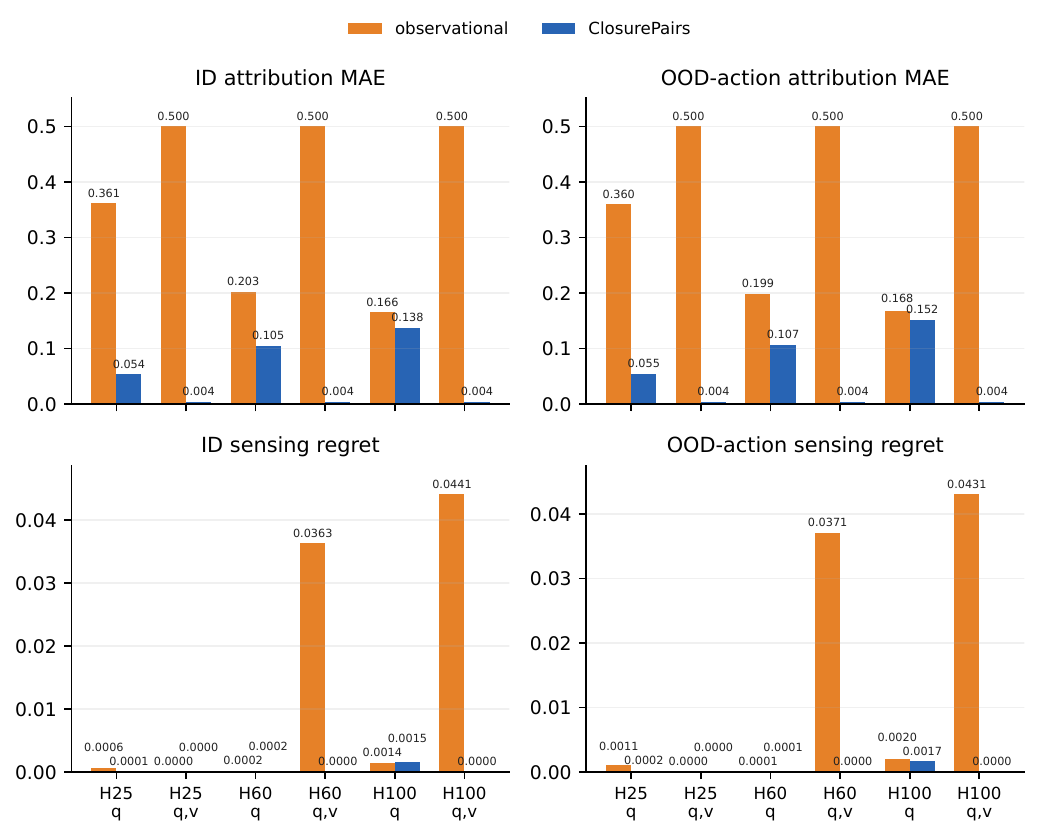}
  \caption{\textbf{Nonlinear multi-step results.} Orange and blue models have
  identical predictive parameters and NLL. Paired supervision recovers the
  mechanism and reduces sensing regret under ID and OOD actions. Each bar
  averages three seeds; $q,v$ denotes full-state observation.}
  \label{fig:nonlinear}
\end{figure}

The five-component MDN predicts final position at horizons $25,60,100$ under
position-only and full-state observations. Training actions lie in
$[-0.25,0.25]$; OOD evaluation uses only $|a|>0.30$ up to $0.55$.
Figure~\ref{fig:nonlinear} gives all matched conditions. The most diagnostic
negative control is full state at horizon 100: the observational split refines
in $88.2\%$ of ID groups (accuracy $0.118$), whereas ClosurePairs never refines
and reaches accuracy $1.0$.

\begin{table}[h]
\caption{Nonlinear matched-run summary. Brackets are bootstrap 95\% intervals
for observational-minus-paired improvement.}
\label{tab:nonlinear}
\centering
\small
\begin{tabular}{llrrr}
\toprule
Split & Method & Fraction MAE & Alias MAE & Decision regret \\
\midrule
ID & Observational & 0.3716 & 0.1000 & 0.01377 \\
ID & ClosurePairs & \textbf{0.0515} & \textbf{0.0439} & \textbf{0.00030} \\
OOD & Observational & 0.3711 & 0.0979 & 0.01390 \\
OOD & ClosurePairs & \textbf{0.0544} & \textbf{0.0453} & \textbf{0.00032} \\
\midrule
ID improvement & 95\% CI & [0.228,0.406] & [0.029,0.088] & [0.0047,0.0226] \\
OOD improvement & 95\% CI & [0.221,0.405] & [0.025,0.084] & [0.0048,0.0226] \\
\bottomrule
\end{tabular}
\end{table}

\subsection{Matched-variance intervention-side routing}
\label{app:matched-routing}

We construct two 12-step stochastic-ball mechanisms. A single frame aliases
velocity in the state mechanism, whereas a future random impulse creates two
modes in the noise mechanism. Both routers receive the same $4\times4$ paired
future matrix and select between the same Gaussian REFINE and five-component
BRANCH heads. The noise-to-state total-variance ratio is $1.021\pm0.021$ over
five seeds. Table~\ref{tab:refine-branch} is therefore a controlled sanity check
that the estimand has the intended decision consequence, not a deployable
online-routing result.

\begin{table}[h]
\caption{Matched-total-variance routing over five seeds (mean $\pm$ standard
deviation). Lower NLL is better.}
\label{tab:refine-branch}
\centering
\small
\begin{tabular}{lrr}
\toprule
Router & Accuracy (\%) & Selected NLL / coordinate \\
\midrule
Always REFINE & $50.00\pm0.00$ & $-1.416\pm0.116$ \\
Always BRANCH & $50.00\pm0.00$ & $-1.767\pm0.019$ \\
Total-variance threshold & $66.48\pm1.06$ & $-2.087\pm0.077$ \\
ClosurePairs & $\mathbf{99.99\pm0.02}$ & $\mathbf{-2.717\pm0.117}$ \\
\bottomrule
\end{tabular}
\end{table}

\subsection{Pixel recurrent details}
\label{app:pixel}

The pixel experiment renders Pendulum as deterministic $32\times32$ grayscale
frames; compatible microstates share pixels but differ in hidden velocity. A
CNN encodes the image, an action-conditioned GRU performs eight latent rollout
steps, and a five-component MDN emits final $\cos\theta$. The ClosurePairs probe
reads the frozen recurrent state. Against the ensemble, its fraction-MAE
improvements are $0.4536$ ID (95\% seed-bootstrap CI
$[0.4086,0.4986]$) and $0.4603$ OOD ($[0.4167,0.5060]$);
alias-variance improvements are $0.02662$ and $0.01936$. The active-sensing
regret difference is $0.00689$ ($[0.00092,0.01376]$), with two-sided Wilcoxon
$p=0.0977$. A separate torque-planning result is inconclusive.

\begin{figure}[h]
  \centering
  \includegraphics[width=0.72\linewidth]{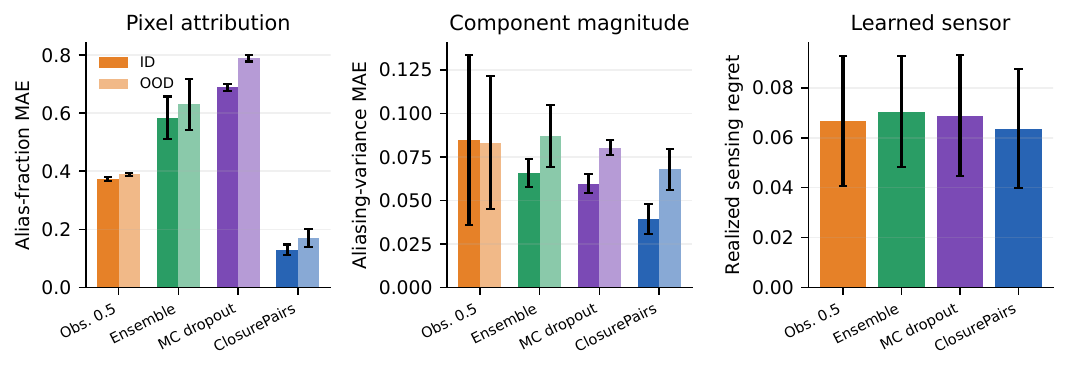}
  \caption{\textbf{Pixel recurrent model.} Fraction and alias-variance MAE use
  crossed targets; sensing regret uses realized forecast loss plus sensor cost.
  Bars show mean $\pm$ standard deviation over ten seeds.}
  \label{fig:pixel}
\end{figure}

\subsection{Pendulum closure curve}
\label{app:pendulum}

We follow Pendulum-v1, add declared Gaussian torque noise, hide angular velocity
at coarse resolution, and predict $\cos\theta$ after 10--100 steps. The
angle-only curve is shown below. At full state, aliasing is zero and process
totals are $0.00085,0.01395,0.04451,0.10202$.

\begin{figure}[h]
  \centering
  \includegraphics[width=0.58\linewidth]{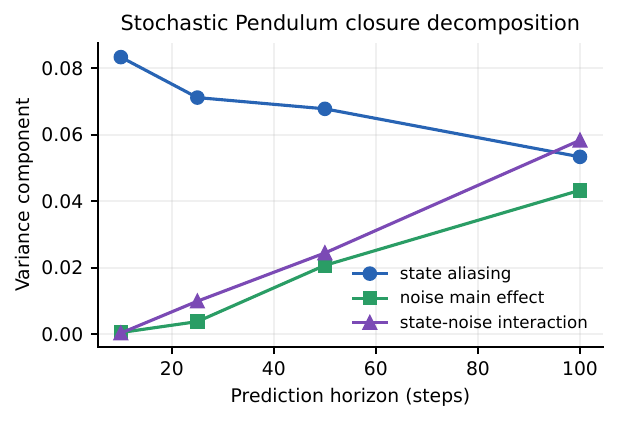}
  \caption{\textbf{Pendulum components.} Hidden-velocity aliasing falls with
  horizon for the bounded outcome, while noise and interaction grow.}
  \label{fig:pendulum}
\end{figure}

\begin{table}[h]
\caption{Stochastic Pendulum crossed decomposition (mean over ten seeds).}
\centering
\small
\begin{tabular}{rrrrr}
\toprule
Horizon & Aliasing & Noise main & Interaction & Process total \\
\midrule
10  & 0.08337 & 0.00051 & 0.00033 & 0.00085 \\
25  & 0.07120 & 0.00380 & 0.00993 & 0.01373 \\
50  & 0.06782 & 0.02072 & 0.02444 & 0.04516 \\
100 & 0.05337 & 0.04331 & 0.05836 & 0.10168 \\
\bottomrule
\end{tabular}
\end{table}

On the learned stochastic Pendulum experiment (horizons 25 and 100, angle-only
and full-state observations, three seeds), pooled observational versus paired
results are: fraction MAE $0.41015$ versus $0.06563$, alias-variance MAE
$0.08638$ versus $0.03747$, decision accuracy $0.50781$ versus $0.84063$, and
decision regret $0.00877$ versus $0.00206$. NLL is identical ($0.08771$). On the
angle-only subsets, fraction attribution improves but decision regret does not
consistently improve because local total-variance error dominates. This is a
failure of the predictive variance magnitude, not of the paired split, and is a
reason to report both calibration and attribution.

\begin{table}[h]
\caption{Pixel recurrent attribution over ten seeds (mean $\pm$ standard
deviation). CP and the observational row share an exactly frozen predictor.}
\centering
\small
\begin{tabular}{llrrr}
\toprule
Split & Method & Fraction MAE & Alias MAE & NLL \\
\midrule
ID & Observational & $0.3718\pm0.0067$ & $0.0846\pm0.0489$ & $-0.1343$ \\
ID & Ensemble & $0.5840\pm0.0728$ & $0.0659\pm0.0081$ & $-0.1640$ \\
ID & MC dropout & $0.6885\pm0.0123$ & $0.0597\pm0.0056$ & $-0.1138$ \\
ID & CP & $\mathbf{0.1304\pm0.0175}$ & $\mathbf{0.0393\pm0.0084}$ & $-0.1343$ \\
OOD & Observational & $0.3894\pm0.0057$ & $0.0832\pm0.0381$ & $0.3057$ \\
OOD & Ensemble & $0.6302\pm0.0882$ & $0.0872\pm0.0178$ & $0.3281$ \\
OOD & MC dropout & $0.7886\pm0.0112$ & $0.0803\pm0.0045$ & $0.3229$ \\
OOD & CP & $\mathbf{0.1699\pm0.0301}$ & $\mathbf{0.0678\pm0.0119}$ & $0.3057$ \\
\bottomrule
\end{tabular}
\end{table}

The active-sensing regrets for never query, always query, observational,
ensemble, MC dropout, and ClosurePairs are respectively $0.12019$, $0.17334$,
$0.06667$, $0.07043$, $0.06884$, and $0.06354$. In the separate torque-planning
stress test, ensemble and ClosurePairs regrets are $0.07154$ and $0.05989$; the
matched improvement CI crosses zero ($[-0.01158,0.04241]$) and the two-sided
Wilcoxon test is not significant ($p=0.652$). The result therefore does not
support a control-improvement claim.

\section{Protocol assumptions and diagnostics}
\label{app:assumptions}

\begin{itemize}
  \item \textbf{Fiber validity:} verify that all $X_i$ map to the same declared
  observation and report the fiber sampler or importance weights.
  \item \textbf{Reset validity:} run deterministic full-state repeats. Any
  positive variance diagnoses reset drift or hidden simulator state.
  \item \textbf{Randomness validity:} distinguish independent nested trials from
  replayable crossed seeds. Never apply the naive shared-noise row estimator to
  a nonlinear system.
  \item \textbf{Resolution validity:} observations must be nested for the
  monotonicity claim; compare the same microstate/action population.
  \item \textbf{Finite samples:} report raw estimates and uncertainty intervals.
  Clipping and isotonic display projections must not be used to manufacture a
  hypothesis-test result.
\end{itemize}

\end{document}